\documentclass[letterpaper, 10 pt, conference]{ieeeconf}  

\IEEEoverridecommandlockouts                              

\usepackage{microtype}
\usepackage{amsmath,amssymb,amsfonts}
\usepackage{graphicx}
\usepackage{textcomp}
\usepackage{xcolor}
\usepackage{subfig}
\usepackage{gensymb}
\usepackage{caption}
\usepackage{hyperref}
\usepackage{mwe}
\usepackage{amsfonts}
\usepackage{amssymb}
\usepackage{verbatim}
\usepackage{amsbsy}
\usepackage{siunitx}
\usepackage{tabularx, booktabs}
\usepackage{dblfloatfix}
\usepackage{cite}
\usepackage{cleveref}

\usepackage[autolanguage]{numprint}

\usepackage{spreadtab}
\usepackage{multirow}
\usepackage{cite}

\usepackage{hyperref}
\usepackage{xcolor}
\usepackage{textcomp}
\usepackage{xcolor}
\usepackage{subfig}
\usepackage{gensymb}
\usepackage{caption}

\usepackage{mwe}
\usepackage{amsfonts}
\usepackage{amssymb}
\usepackage{verbatim}
\usepackage{amsbsy}
\usepackage{siunitx}
\usepackage{tabularx, booktabs}
\usepackage{soul}
\usepackage{tikz}
\usetikzlibrary{calc}
\usepackage{algorithm}
\usepackage{algpseudocode}
\usepackage{amsmath}

\makeatletter
\newif\if@anonymize

\@anonymizetrue    

\if@anonymize
  \newcommand{\highlight@DoHighlight}{
    \fill [outer sep = -15pt, inner sep = 0pt, color=black]
          ($(begin highlight)+(0,8pt)$) rectangle ($(end highlight)+(0,-3pt)$) ;
  }

  \newcommand{\highlight@BeginHighlight}{
    \coordinate (begin highlight) at (0,0) ;
  }

  \newcommand{\highlight@EndHighlight}{
    \coordinate (end highlight) at (0,0) ;
  }

  \newdimen\highlight@previous
  \newdimen\highlight@current
  \newlength{\item@width}

  \DeclareRobustCommand*\anonymize{%
    \SOUL@setup
    \def\SOUL@preamble{%
      \begin{tikzpicture}[overlay, remember picture]
        \highlight@BeginHighlight
        \highlight@EndHighlight
      \end{tikzpicture}%
    }%
    \def\SOUL@postamble{%
      \begin{tikzpicture}[overlay, remember picture]
        \highlight@EndHighlight
        \highlight@DoHighlight
      \end{tikzpicture}%
    }%
    \def\SOUL@everyhyphen{%
      \discretionary{%
        \SOUL@setkern\SOUL@hyphkern
        \SOUL@sethyphenchar
        \tikz[overlay, remember picture] \highlight@EndHighlight ;%
      }{%
      }{%
        \SOUL@setkern\SOUL@charkern
      }%
    }%
    \def\SOUL@everyexhyphen##1{%
      \SOUL@setkern\SOUL@hyphkern
      \settowidth{\item@width}{##1}%
      \makebox[\item@width]{}%
      \discretionary{%
        \tikz[overlay, remember picture] \highlight@EndHighlight ;%
      }{%
      }{%
        \SOUL@setkern\SOUL@charkern
      }%
    }%
    \def\SOUL@everysyllable{%
      \begin{tikzpicture}[overlay, remember picture]
        \path let \p0 = (begin highlight), \p1 = (0,0) in \pgfextra
          \global\highlight@previous=\y0
          \global\highlight@current =\y1
        \endpgfextra (0,0) ;
        \ifdim\highlight@current < \highlight@previous
          \highlight@DoHighlight
          \highlight@BeginHighlight
        \fi
      \end{tikzpicture}%
      \settowidth{\item@width}{\the\SOUL@syllable}%
      \makebox[\item@width]{}%
      \tikz[overlay, remember picture] \highlight@EndHighlight ;%
    }%
    \SOUL@
  }
\else
  \newcommand{\anonymize}[1]{#1}
\fi

\newcommand{\linebreakand}{%
  \end{@IEEEauthorhalign}
  \hfill\mbox{}\par
  \mbox{}\hfill\begin{@IEEEauthorhalign}
}

\makeatother

\title{\LARGE \bf
Cross domain Persistent Monitoring for \\ Hybrid Aerial Underwater Vehicles
}

\author{Ricardo B. Grando$^{1,3}$, Victor A. Kich$^{2}$, Alisson H. Kolling$^{1}$,\\Junior C. D. Jesus$^{1}$, Rodrigo S. Guerra$^{1}$, Paulo L. J. Drews-Jr$^{1}$
}

\begin{document}

\maketitle
\thispagestyle{empty}
\pagestyle{empty}






\maketitle
\thispagestyle{empty}
\pagestyle{empty}

\begin{abstract}

Hybrid Unmanned Aerial Underwater Vehicles (HUAUVs) have emerged as platforms capable of operating in both aerial and underwater environments, enabling applications such as inspection, mapping, search, and rescue in challenging scenarios. However, the development of novel methodologies poses significant challenges due to the distinct dynamics and constraints of the air and water domains. In this work, we present persistent monitoring tasks for HUAUVs by combining Deep Reinforcement Learning (DRL) and Transfer Learning to enable cross-domain adaptability. Our approach employs a shared DRL architecture trained on Lidar sensor data (on air) and Sonar data (underwater), demonstrating the feasibility of a unified policy for both environments. We further show that the methodology presents promising results, taking into account the uncertainty of the environment and the dynamics of multiple mobile targets. The proposed framework lays the groundwork for scalable autonomous persistent monitoring solutions based on DRL for hybrid aerial-underwater vehicles.

\end{abstract}


\section{Introduction}
\label{introduction}

Hybrid Unmanned Aerial Underwater Vehicles (HUAUVs) have recently been a topic of research \cite{pinheiro2024energy, costa2025image, dias4992978hydrone, bedin2021deep}, given their unique set of tasks that leverage their air and water characteristics to operate effectively in both environments. They can help, for example, in improving inspection and mapping in oil platforms and submerged industrial facilities, besides allowing for better search and rescue tasks in harsh scenarios. However, autonomous operation in two different environments includes additional problems, such as wind and energy autonomy in the air and current and turbidity while underwater. These limitations have been incentivizing the research of novel approaches to improve the vehicle's capability to perform basic tasks such as navigation \cite{bedin2021deep, de2022depth, grando2022mapless}.

\begin{figure*}[tbp!]
    \centering
    \subfloat[First environment without obstacles, representing the baseline scenario for training the HUAUV’s persistent monitoring policy.]{%
        \begin{minipage}[c]{0.22\linewidth}
            \centering
            \includegraphics[width=\linewidth]{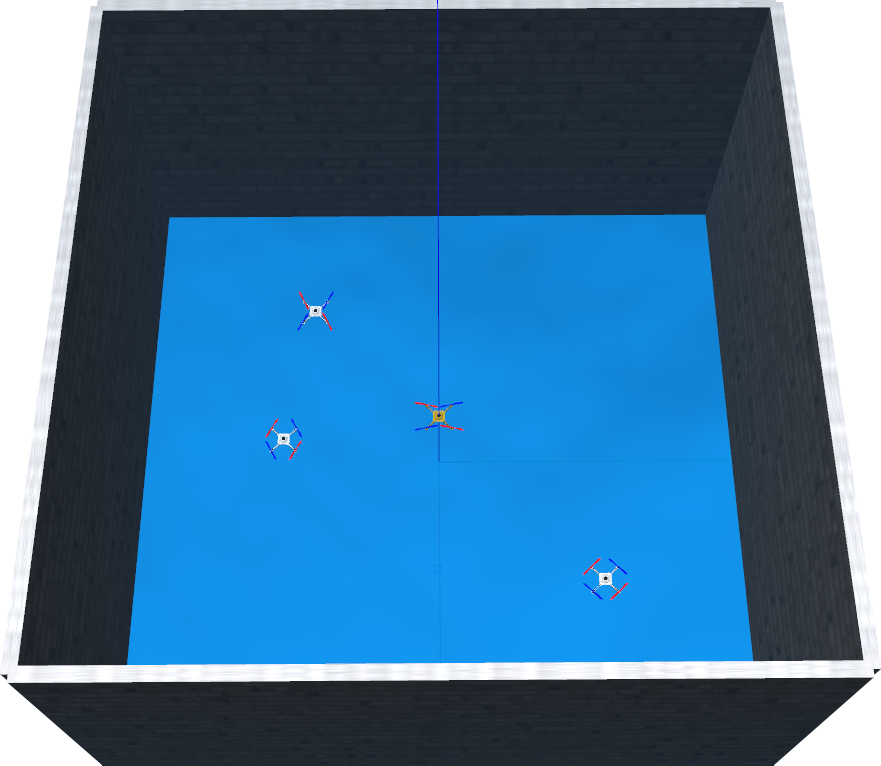}%
            \label{fig:img1}
        \end{minipage}
    }
    \hfill
    \subfloat[Second environment with four cylinders simulating drilling risers, used to evaluate obstacle-avoidance behavior.]{%
        \begin{minipage}[c]{0.22\linewidth}
            \centering
            \includegraphics[width=0.87\linewidth]{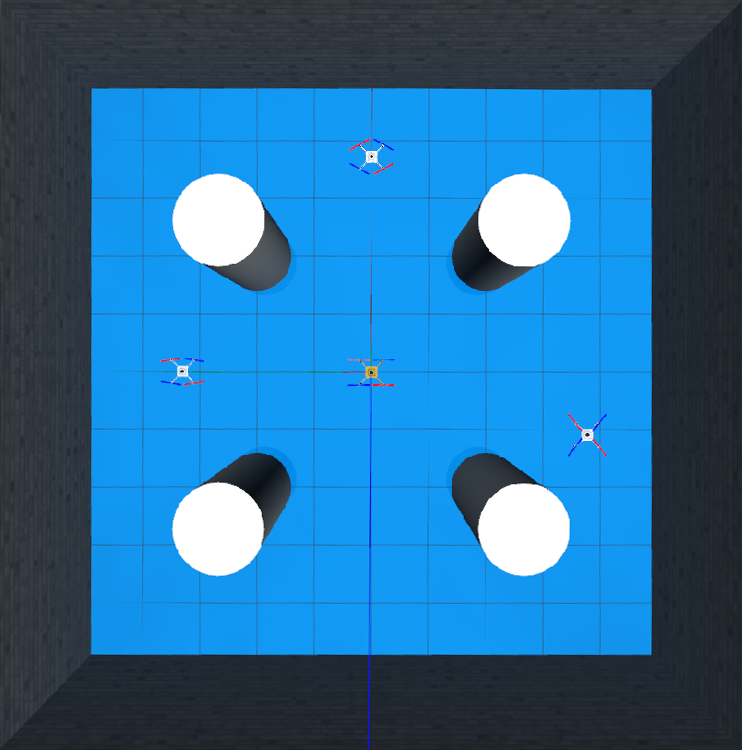}%
            \label{fig:img2}
        \end{minipage}
    }
    \hfill
    \subfloat[Visualization of the HUAUV agents in the obstacle-free  environment, showing coverage cones and coordination for target uncertainty reduction.]{%
        \begin{minipage}[c]{0.22\linewidth}
            \centering
            \includegraphics[width=\linewidth]{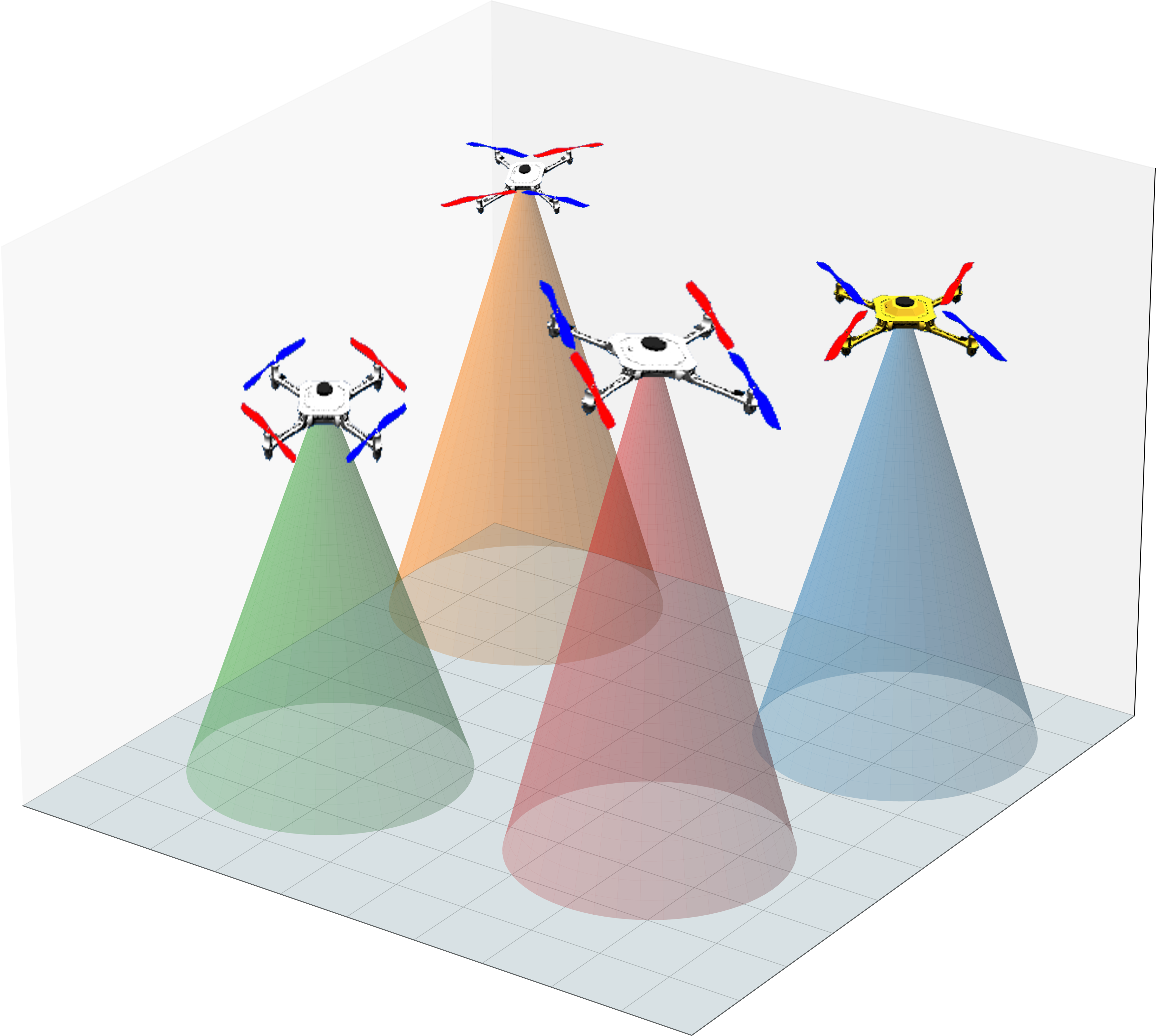}%
            \label{fig:img3}
        \end{minipage}
    }
    \hfill
    \subfloat[Visualization of the HUAUV agents in the environment with cylindrical obstacles, illustrating simultaneous coverage and obstacle avoidance.]{%
        \begin{minipage}[c]{0.22\linewidth}
            \centering
            \includegraphics[width=\linewidth]{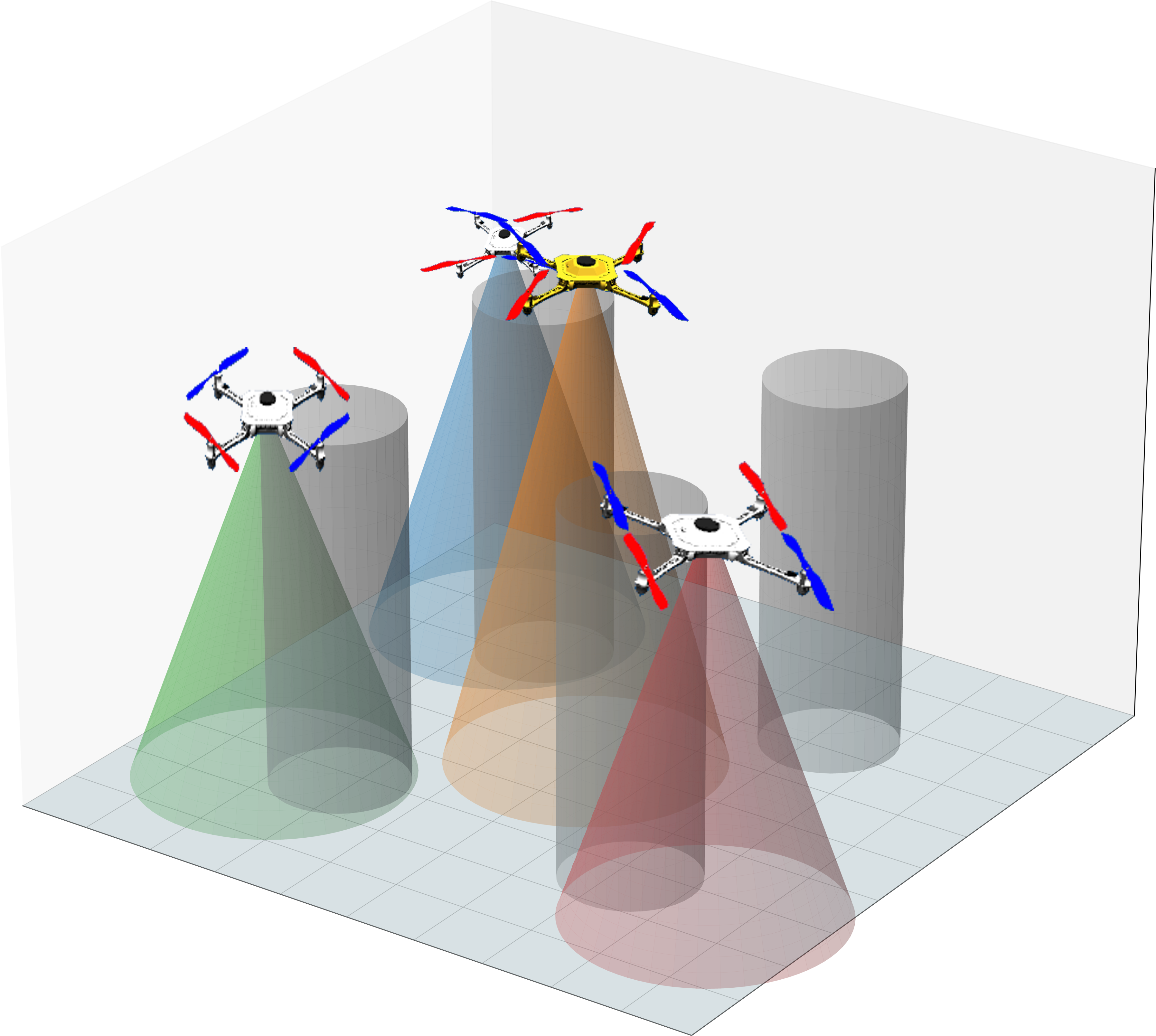}%
            \label{fig:img4}
        \end{minipage}
    }
    \caption{Simulated HUAUV persistent monitoring environments. (a) Aerial domain, Env~1 (obstacle-free). (b) Aerial domain, Env~2 (four cylindrical obstacles). (c)--(d) Example LiDAR scans in Env~1 and Env~2, respectively. Underwater scenarios use the same layouts but with sonar range measurements.}
    \label{fig:environments}
    \vspace{-6.5mm}
\end{figure*}

Several approaches have already been proposed for autonomous navigation of mobile robots based on traditional algorithms and, more recently, based on Deep Reinforcement Learning (DRL) \cite{zhu2021deep}. However, for more complex tasks such as the persistent monitoring task, there is still a need for scientific contribution \cite{basilico2022recent}. Specifically, considering a multimedia context such as air-water, cross-domain methodologies can be a way forward to enable this kind of vehicle to perform such tasks, improving its overall ability to be aware of its surroundings and providing scalability on its use for tasks not only in monitoring, but also in search and rescue and other related tasks. Persistent monitoring consists of tasks related to the continuous or repeated observation of an area or a set of targets. Also, since DRL algorithms have been successfully used for autonomous navigation-related tasks for mobile robots in general, this task could benefit from the methodology’s ability to handle the additional complexity it may demand.

In this work, we introduce persistent monitoring tasks for HUAUV based on DRL. Considering that the vehicle needs to act in two different environments, we propose a methodology that can be applied in both environments independently of the differences between them. For that, we address the problem based on a ranging sensor point of view, considering a vehicle that has a Lidar for aerial sensing and a Sonar for underwater perception. Based on this kind of sensing, we demonstrate that DRL-agents with the same architecture can be transfer-learned to act in a different environment, providing a suitable monitoring capability while avoiding the use of visual or other positional information. We base our approach on a Parallel DRL architecture and on Transfer Learning technique, where the proof of concept based a single HUAUV and multiple moving targets can be improved and scaled up to more complex scenarios. Overall, our work provides the following contributions:

\begin{itemize}

\item We present a proof-of-concept methodology for persistent monitoring tasks with Hybrid Unmanned Aerial Underwater Vehicles (HUAUVs) based on Deep Reinforcement Learning (DRL) and Transfer Learning, showing that a single policy can be deployed across aerial and underwater domains using LiDAR and sonar range data.

\item We propose a distributed parallel DRL training setup that accelerates policy learning and we show, in simulation, that the learned policy can manage the average uncertainty of multiple mobile targets more efficiently than a Bug2-style behaviour-based baseline, both in the original aerial domain and after zero-shot transfer to the underwater domain.

\item We systematically analyse a simplified 2D setting with realistic vehicle dynamics and sensing, and we discuss the limitations of this formulation and directions towards more complex 3D scenarios and richer persistent monitoring objectives.

\end{itemize}

This work has the following structure: the related works are discussed in the following Section (Sec. \ref{related_works}). We present our methodology in Sec. \ref{methodology} and the results in Sec. \ref{results}. Finally, we present our contributions and future works in Sec. \ref{conclusion}.

\section{Related Work} 
\label{related_works}

Persistent Monitoring tasks for mobile robots are still focused mostly on Unmanned Ground Vehicles (UGVs) \cite{basilico2022recent}. As for HUAUVs, the literature is still mostly concerned with mechanical design and modeling \cite{drews2014hybrid, neto2015attitude, da2018comparative, maia2017design, lu2019multimodal, mercado2019aerial, horn20}. A few DRL methodologies have been already proposed for navigation in HUAUVs \cite{bedin2021deep, grando2022mapless} and for persistent monitoring with mobile robots \cite{kurunathan2021deep}, \cite{asarkaya2021temporal}, \cite{wang2023spatio}, \cite{sun2023persistent}, \cite{mishra2024deep}, \cite{mishra2024multi}, \cite{vashisth2024deep}.

Want et al. \cite{wang2023spatio} introduced a spatiotemporal attention ANN built on \textit{Transformer}-style attention layers that processes the agent's belief and the graph-structured environment to generate next landmark decisions. The problem was formulated as a UAV navigating in a 2D environment that must track moving ground targets. The vehicle was equipped with an RGB camera with a limited field of view. The metrics used to validate the models were visit count, estimation error, and target uncertainty. The methodology was compared with non-DRL-based algorithms. Overall, this work presents a DRL-based attention model based on visual information that learns when and where to navigate to reduce uncertainty about moving targets.

Mishra et al. \cite{mishra2024deep} presented a DRL-based framework for managing persistent surveillance missions with energy-constrained UAVs. The task is to determine an optimal sequence of visits to a set of targets by a single UAV, minimizing the maximum time elapsed between successive visits to any target, while ensuring that the UAV does not run out of fuel or battery. The UAV is initially parked at a depot and must return periodically for refueling or battery recharge due to fuel or flight time constraints. A UAV was used, and the model was trained using the PPO algorithm with information related to the vehicle's position. The UAV is tasked with repeatedly visiting a set of targets with equal priority, ensuring persistent surveillance. Due to limited fuel or battery life, the UAV must return to the depot for refueling or recharge before depleting its energy reserves. In this way, the UAV seeks to minimize the maximum revisit time for all targets, ensuring that no target remains unobserved for long periods. The average number of visits was used to validate the methodology.


Sun et al. \cite{sun2023persistent} presented a DRL-based framework for persistent coverage tasks using multiple UAVs. The UAVs are tasked with persistently covering a target region, ensuring that the area is continuously monitored over time. Each UAV uses the \textit{Wonderful Life Utility} (WLU) function to evaluate the impact of its actions on overall system performance, fostering decisions that improve collective coverage efficiency. The WLU functions are used to guide the UAVs in making decisions that consider individual and collective benefits, promoting cooperative behavior. A total of eight UAVs were used to validate the methodology, trained following the actor-critic DRL algorithm based on the UAVs' position. The metrics used to validate the methodology were the rewards obtained during training, coverage, and navigation examples.


Vashisth et al. \cite{vashisth2024deep} demonstrated an adaptive DRL approach for persistent image-based tracking. The UAV seeks to find targets, avoid obstacles, and adapt to truly unknown environments in real time. The methodology employs an actor-critic neural network within the policy, trained to select landmarks using dynamic, locally sampled landmark graphs. A multi-rotor UAV equipped with a ground-pointing RGB-D camera was used. At each time step, candidate landmarks in the UAV's local free space are defined. Afterward, a dynamic graph of reachable waypoints and yaw combinations is constructed. Finally, the RL algorithm is used to select the next landmark. The methodology was tested in a photorealistic orchard environment using AirSim + Unreal Engine. Example trajectories and object detection rates were used to validate the methodology.


Our work differs from the previously discussed works by focusing on a cross-domain methodology with transfer learning, where the persistent monitoring task is assigned for the aerial and underwater domains. We also propose a novel methodology based on distributed parallel DRL, which provides faster learning and a scalable methodology for more complex scenarios when compared to non-parallel methods. Our methodology is also based on information from a ranging distance sensor to perform the persistent monitoring task and avoid obstacles, which present fewer limitations in both aerial and underwater domains when compared with image-based or GPS-based approaches.

\section{Methodology} 
\label{methodology}

In this section, we discuss the proposed persistent monitoring problem and our proposed DRL methodology. We detail the monitoring system, method, and the simulation setup. 

\subsection{Problem Setup}

We consider a persistent monitoring task in bounded 2D scenarios \( \mathcal{E} \subset \mathbb{R}^2 \), where a single autonomous agent, modeled as a continuous-control DRL policy, must reduce and manage the uncertainty of \( N \) mobile targets. Each target \( i \) moves according to a predefined Lissajous curve path. The physical HUAUV is simulated with full 6-DoF dynamics, but for the DRL formulation we use a 2D kinematic state:


\[
s_t = [x_t, y_t]^\top, \quad a_t = [v_t, \omega_t]^\top
\]
where \( v_t \) and \( \omega_t \) are linear and angular velocity setpoints forwarded to the low-level controllers of the RotorS (air) and UUV (water) simulators. The resulting state evolution can be written abstractly as

\[
s_{t+1} = s_t + \Delta t \cdot f(s_t, a_t)
\]
where \( f(\cdot) \) captures the closed-loop dynamics induced by the underlying vehicle model, controllers, and disturbances. 

In the aerial and underwater domains we obtain different functions \(f_{\text{air}}\) and \(f_{\text{water}}\) because of the different vehicle dynamics and control loops. Thus, although the RL state is 2D, the policy experiences a genuine domain shift in its transition dynamics when transferred from the aerial to the underwater simulator.

\subsection{Observations and Target Uncertainty}

The agent observes the current 2D state and noisy estimates of each target’s position:
\begin{equation}
    o_t = \{ s_t, \hat{g}_{1,t}, \hat{g}_{2,t}, \hat{g}_{3,t} \},
\end{equation}
where \(\hat{g}_{i,t} \in E\) is the estimated position of target \(i\) at time \(t\), obtained from range observations.

Each target \(i\) has an associated uncertainty \(\sigma_{i,t} \in \mathbb{R}_+\) that evolves as
\begin{equation}
    \sigma_{i,t+1} =
    \begin{cases}
        0, & \text{if } \lVert s_t - g_{i,t} \rVert \le r_{\text{sense}}, \\
        \sigma_{i,t} + \lambda \, \Delta t, & \text{otherwise},
    \end{cases}
\end{equation}
where \(g_{i,t}\) is the true target position, \(r_{\text{sense}}\) is the sensing radius, and \(\lambda > 0\) is the linear uncertainty growth rate.

The policy \( \pi_\phi(a_t|o_t) \) is parameterized by a fully connected neural network trained using the DSAC algorithm. When resetting the uncertainty, the agent receives a positive reward \( r \). This encourages the agent to reduce the uncertainty of all targets as efficiently and frequently as possible.

The proposed DSAC-based policy is trained in simulation operating in the aerial domain. The DSAC policy is trained in a 2D aerial environment \( \mathcal{E}_{\text{air}} \subset \mathbb{R}^2 \) where the HUAUV exhibits relatively fast, low-drag dynamics. After training, the learned policy \( \pi_\phi \) is transferred to the underwater domain \( \mathcal{E}_{\text{water}} \subset \mathbb{R}^2 \). The underwater dynamics differ due to hydrodynamic drag and slower accelerations:
\[
s_{t+1}^{\text{water}} = s_t + \Delta t \cdot f_{\text{water}}(s_t, a_t)
\]
with \( f_{\text{water}}(\cdot) \neq f_{\text{air}}(\cdot) \), representing a domain shift.

The same state and action spaces are shared across domains: \( \mathcal{S}_{\text{air}} = \mathcal{S}_{\text{water}} \), \( \mathcal{A}_{\text{air}} = \mathcal{A}_{\text{water}} \). The policy \( \pi_\phi \) trained via DSAC in the aerial domain is deployed in the underwater domain without retraining (zero-shot transfer). The reward structure and target uncertainty model remain unchanged.

\subsection{Distributional Soft Actor–Critic (DSAC)}

The Distributional Soft Actor–Critic (DSAC), proposed by \cite{ma2020dsac}, combines the maximum entropy reinforcement learning framework of SAC with principles from distributional reinforcement learning. Instead of learning scalar Q-values,

\begin{equation}
    Q_\theta(s, a) \approx \mathbb{E}\left[ \sum_{t=0}^\infty \gamma^t r_t \right],
\end{equation}

\noindent DSAC models the full distribution over returns using parametric distributions, the Categorical distribution in our work,

\begin{equation}
    Z_\theta(s, a) \approx \text{Cat}\Biggl(\sum_{t=0}^\infty \gamma^t r_t \Biggr).
\end{equation}

\noindent Thereby, capturing uncertainty in value estimates more effectively. This approach enhances both the stability and performance of the learning process, especially in complex continuous control tasks.

Standard SAC estimates the expected value of return via scalar Q-functions, which can be sensitive to bootstrapping errors. DSAC, by contrast, learns the full return distribution using distributional critics, allowing better uncertainty modeling, reduced overestimation bias, and more stable policy gradients. DSAC is built on the same maximum entropy principle as SAC, encouraging stochastic exploration by maximizing both expected reward and policy entropy. DSAC maintains a stochastic policy \( \pi(a|s;\phi) \), modeled as a Gaussian distribution (actor), and a pair of distributional Q-networks \( Z_1(s, a;\theta_1) \), \( Z_2(s, a;\theta_2) \), where each outputs the quantiles of a return Categorical distribution. It also maintains a corresponding set of target Q-networks \(Z_{\bar{\theta_{1,2}}}\) and soft target updates. 

We parameterise the return distribution \(Z_\theta(s,a)\) as \(N\) quantiles \(\{ z_i(s,a) \}_{i=1}^N\) with associated quantile fractions \(\{ \tau_i \}_{i=1}^N\), where \(\tau_i \in (0,1)\) and \(\sum_i \tau_i = 1\). In all experiments we use \(N = 64\) quantiles. Here, \(\tau_i\) denotes the quantile fraction of the \(i\)-th critic output and is \emph{not} related to the target uncertainty \(\sigma_{i,t}\) defined in the monitoring problem. 

\begin{table*}[b]
\centering
\setlength{\tabcolsep}{13.5pt}
\caption{Average Time and Uncertainty.}
\label{table:uncertain_table}
\begin{tabular}{c c c c c c c c c} 
\toprule
Model & $t_{air}$ (s) & $t_{water}$ (s) & $\tau_{1 air}$ & $\tau_{2 air}$  & $\tau_{3 air}$ & $\tau_{1 und}$ & $\tau_{2 und}$  & $\tau_{3 und}$  \\
\midrule
\textbf{DSAC (Env1)} & $35.40$ $\pm$ $30.61$ & $\textbf{97.95}$ $\pm$ $\textbf{83.62}$ & $\textbf{9.91}$ & $\textbf{9.66}$ & $ \textbf{9.08}$ & $\textbf{27.75}$ & $\textbf{27.45}$ & $\textbf{23.84}$ \\
\textbf{BUG2 (Env1)} & $\textbf{36.90}$ $\pm$ $\textbf{16.98}$ & $127.43$ $\pm$ $56.85$ & $10.32$ & $10.20$ & $12.51$ & $35.54$ & $36.69$ & $42.12$ \\
\textbf{DSAC (Env2)} & $\textbf{49.34}$ $\pm$ $\textbf{49.45}$ & $\textbf{133.67}$ $\pm$ $\textbf{142.21}$ & $\textbf{12.98}$ & $\textbf{10.41}$ & $\textbf{13.39}$ & $\textbf{26.62}$ & $\textbf{32.87}$ & $\textbf{27.72}$ \\
\textbf{BUG2 (Env2)} & $93.63$ $\pm$ $119.55$ & $155.73$ $\pm$ $151.44$ & $30.06$ & $37.82$ & $26.09$ & $36.92$ & $45.20$ & $35.27$ \\

\bottomrule
\end{tabular}
\end{table*}

The learning process is based on the distributional Bellman equation:
\begin{equation}
Z(s, a) \overset{D}{=} r(s, a) + \gamma \, Z(s', a'), \quad a' \sim \pi_\phi(\cdot | s'),
\end{equation}
where $\overset{D}{=}$ denotes equality in distribution. 

In practice, $Z_\theta(s, a)$ is parameterized as a finite set of quantiles $\{ z_i(s, a) \}_{i=1}^N$. For each sampled transition $(s, a, r, s')$, the target quantiles are computed as:
\begin{multline}
y_j = r(s, a) + \gamma \, z'_j(s', a'), \\ a' \sim \pi_\phi(\cdot | s'), \ z'_j \sim Z_{\bar{\theta}}(s', a'),
\end{multline}
where $Z_{\bar{\theta}}$ denotes the target network for the distributional critic.

The critical parameters are optimized using the Quantile Huber Loss:
\begin{equation}
\mathcal{L}_{Z_{1,2}}(\theta) = \frac{1}{N} \sum_{i=1}^N \sum_{j=1}^N \rho_{\tau_i}^\kappa \left( y_j - z_i(s, a) \right),
\end{equation}
where $\tau_i$ is the quantile fraction and $\rho_{\tau}^\kappa$ is the quantile regression Huber loss defined as:
\begin{multline}
\rho_{\tau}^\kappa(u) = |\tau - \mathbb{I}\{ u < 0 \}| \
\mathcal{L}_\kappa(u), 
\\
\mathcal{L}_\kappa(u) =
\begin{cases}
\frac{1}{2}u^2 & \text{if } |u| \leq \kappa, \\
\kappa (|u| - \frac{1}{2} \kappa) & \text{otherwise}.
\end{cases}
\end{multline}

The actor retains SAC’s entropy-regularized objective, encouraging both high return and policy entropy:
\begin{equation}
J_\pi(\phi) = \mathbb{E}_{s \sim \mathcal{D}, a \sim \pi_\phi} 
\left[ \min_{i} Q_{\theta_i}(s, a) - \alpha \log \pi_\phi(a | s) \right].
\end{equation}
Here, $Q_{\theta_i}$ is obtained as the expectation of the predicted return distribution, $\mathbb{E}[Z_\theta(s, a)]$, although other distributional statistics (e.g., lower quantiles) can be used to incorporate risk-sensitive behavior. The temperature parameter $\alpha$ is adjusted automatically to balance exploration and exploitation.

The DSAC policy is trained only in the aerial domain using LiDAR observations. All underwater results are obtained by deploying the aerially trained policy in the underwater simulator without any further fine-tuning, in a transfer leaning or a zero-shot transfer setting. We do not train a separate underwater policy



\subsection{Simulated Environments and Vehicle Description}

Our simulated environments are built on the Gazebo simulator and integrated with ROS. It makes use of the RotorS framework \cite{furrer2016rotors}, which supports the simulation of aerial vehicles with various control modes, such as angular rates, attitude, and position control. The simulation also includes a wind simulation using an Ornstein-Uhlenbeck noise model. 

Alongside RotorS, the UUV Simulator \cite{manhaes2016uuv} was used for underwater simulation. This framework enables the simulation of hydrostatic and hydrodynamic forces, thrusters, sensors, and external disturbances. Using this framework, the underwater model of the vehicle was configured with parameters including volume, added mass, center of buoyancy, and other relevant physical properties, along with the characteristics of the underwater environment. 

We consider two planar environments \(E_1\) and \(E_2\) in both the aerial and underwater domains. Environment \(E_1\) is an obstacle-free \(10 \,\text{m} \times 10 \,\text{m}\) tank, used as a baseline for learning the monitoring policy. Environment \(E_2\) adds four cylindrical obstacles representing drilling risers. In both environments we evaluate the policy first in the aerial domain (LiDAR) and then in the underwater domain (sonar) without retraining. Figure \ref{fig:environments} highlights the scenarios used to validate the proposed methodology.

The vehicle used in our experiments is based on the model introduced by Drews-Jr et al. \cite{drews2014hybrid}, Neto et al. \cite{neto2015attitude}, and further refined by Horn et al. \cite{horn2019study}. It was modeled according to its real mechanical specifications, including parameters such as inertia, motor coefficients, mass, rotor velocity, among others. The onboard sensing system was configured to reflect real-world LIDAR and sonar capabilities. The simulated LIDAR is based on the Hokuyo UST-10LX model, offering a sensing range of up to 10 meters with a 270° field of view and an angular resolution of 0.25°. It was implemented using Gazebo’s ray-based sensor plugin. For sonar simulation, a forward-looking sonar (FLS) was modeled using the plugin developed by Cerqueira et al. \cite{cerqueira2017novel}. This sonar features a maximum range of 20 meters, with 1000 bins and 256 beams. The beam’s angular dimensions were set to 90° horizontally and 15° vertically. 

The three moving targets used in both scenarios have a predefined Lissajous curve, with a random linear velocity to follow the path. The vehicles follow the paths in a persistent manner, having a linear uncertainty function associated with them and sharing their position with the DRL agent. 

The Lissajous curve has the form
\begin{equation}
    g_{i,t} = \big( A_i \sin(a_i t + \phi_i),\; B_i \sin(b_i t + \psi_i) \big),
\end{equation}
with amplitudes \(A_i, B_i \in [2, 5]~\text{m}\) and integer frequencies \(a_i, b_i\) of 2. For each episode, we sample the phases \(\phi_i, \psi_i\) uniformly in \([0, 2\pi)\) and draw the linear speed along the curve from a uniform distribution \(v_i \sim \mathcal{U}(0.1, 0.25)~\text{m/s}\). 

In all experiments, the DRL agent only observes the reduced 2D state \(s_t\) and range data, while the underlying RotorS and UUV Simulator instances evolve according to the full six-degrees-of-freedom hybrid vehicle model. The air and water dynamics therefore differ in the internal mapping from the commanded setpoints \((v_t, \omega_t)\) to the realised planar motion.


\subsection{Behaviour-Based Baseline (BUG2)}

As a behaviour-based baseline we adopt the minimalistic quadrotor navigation strategy of Marino et al.~\cite{marino2016minimalistic}, which is conceptually similar to a Bug2 algorithm and has been used previously for indoor aerial navigation. In our implementation, the baseline follows Bug2-style obstacle avoidance and target-seeking behaviour using range measurements only. We evaluate it in the same single-agent, multi-target monitoring task as DSAC.


\section{Experimental Results}
\label{results} 

\begin{figure*}[ht]
\vspace{-6.5mm}
  \centering
  \subfloat[Uncertainty through time in the first environment (Air).\label{fig:air_1}]{
	\begin{minipage}[c][0.38\width]{0.23\linewidth}
	   \centering
	   \includegraphics[width=\linewidth]{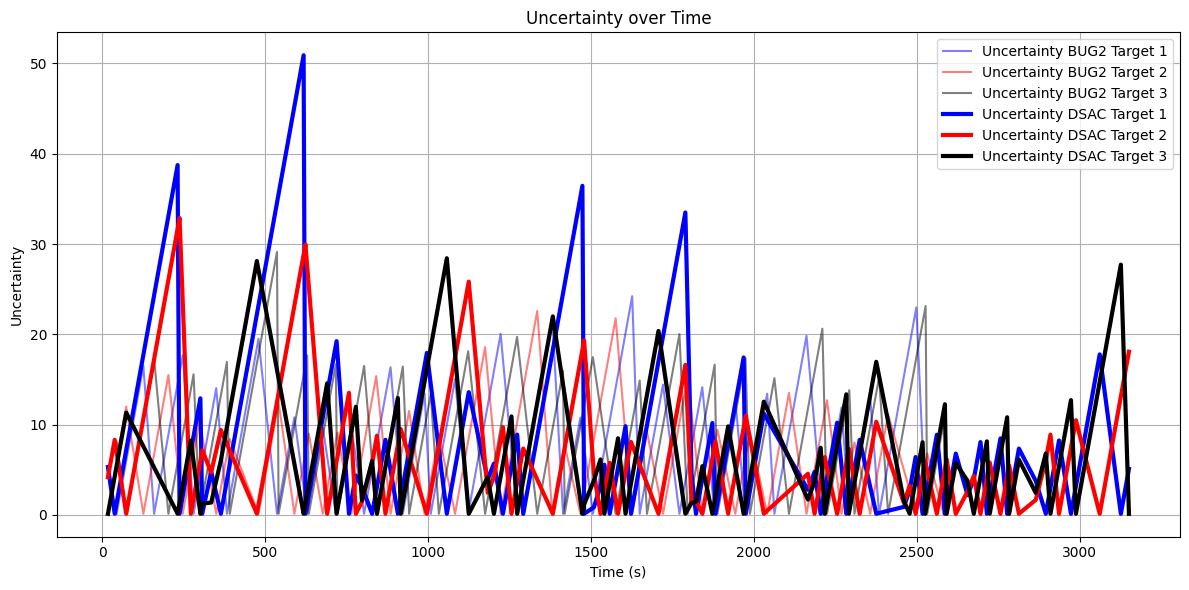}
	\end{minipage}}
 \hfill 	
  \subfloat[Uncertainty through time in the second environment (Air).\label{fig:air_2}]{
	\begin{minipage}[c][0.38\width]{0.23\linewidth}
	   \centering
	   \includegraphics[width=\linewidth]{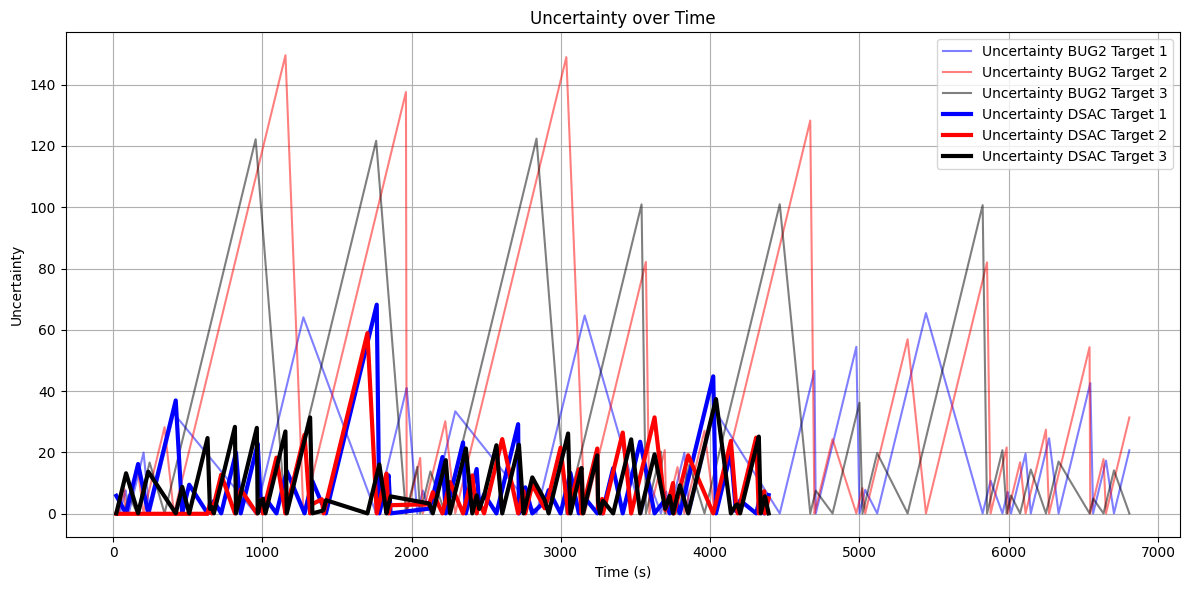}
	\end{minipage}}
   \hfill 	
  \subfloat[Uncertainty through time in the first environment (Underwater).\label{fig:und_1}]{
	\begin{minipage}[c][0.38\width]{0.23\linewidth}
	   \centering
	   \includegraphics[width=\linewidth]{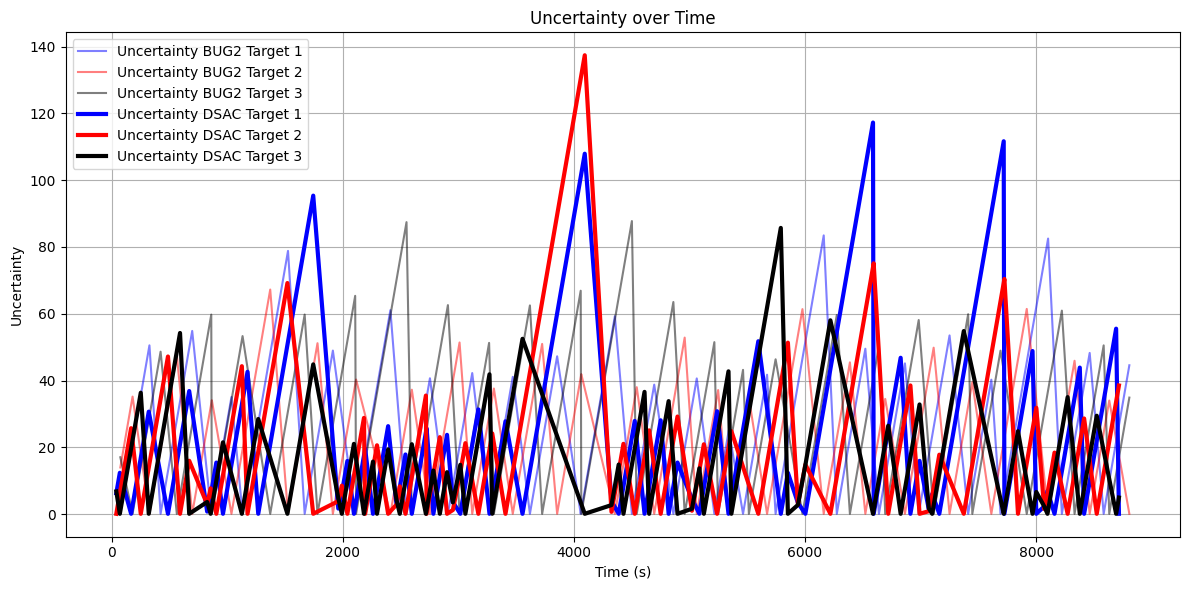}
	\end{minipage}}
  \subfloat[Uncertainty through time in the second environment (Underwater).\label{fig:und_2}]{
	\begin{minipage}[c][0.38\width]{0.23\linewidth}
	   \centering
	   \includegraphics[width=\linewidth]{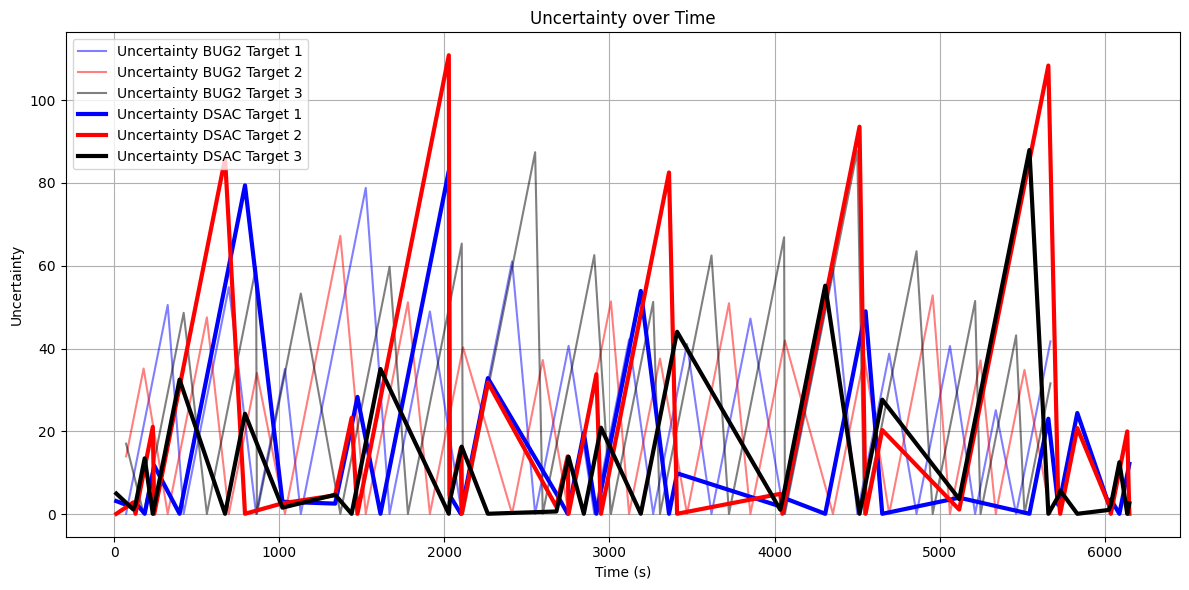}
	\end{minipage}}
\caption{Evolution of target uncertainties \(\sigma_{i,t}\) during evaluation for Env~1 and Env~2 in the aerial and underwater domains. The vertical axis shows the uncertainty, and the horizontal axis is the simulation time steps. DSAC reduces the peaks and frequency of high-uncertainty intervals compared to the Bug2 baseline.}
\label{fig:uncertaintyepisodes}
\end{figure*}



A generated path was created for each moving target \(i\) during the training and validation phase, with a randomly generated velocity at each step for each target \(i\). During training, each episode starts with three targets initialised on their Lissajous trajectories with uncertainties \(\sigma_{i,0} \approx 0\). Whenever the agent enters the sensing radius of a target, the corresponding uncertainty \(\sigma_{i,t}\) is reset to zero, the agent receives a positive reward, and the next target to pursue is selected as the one with highest current uncertainty.

The agents train until they reset at least one of the target's uncertainty or until they collide with an obstacle or with the tank border. A learning rate of $10^{-3}$ was used, alongside a minibatch of 256 samples and the Adam optimizer in the proposed ANN. We limited the number of episodes to 2500, where each episode terminates either when a fixed horizon of \(T = 5000\) simulation steps is reached or when the agent collides. We consider that the \( \lambda > 0 \) linear uncertainty \(\bar{\sigma}_i\) grows by $0.001$ each simulated step, starting with a random value for each target \(i\) close to 0. 

For evaluation, we use the same setting and let the agent run for the full horizon \(T\). We then compute the average uncertainty \(\bar{\sigma}_i\) of each target \(i\) and the average time-to-visit of each target. We denote by \(t_{\text{air}}\) and \(t_{\text{water}}\) the average times to first visit in the aerial and underwater domains, respectively, averaged over 100 trials.

We collected the statistics for our methodology and compared them with the performance of a behavior-based algorithm \cite{marino2016minimalistic}. The evaluation was performed for 100 trials, and the average amount of uncertainty \(\bar{\sigma}_i\)  was recorded. Also, the average time for both underwater ($t\_water$) and aerial ($t\_air$) along navigation with their standard deviations was recorded. Specifically for the underwater evaluation, the evaluation was done using the models trained in the aerial setup. 
Table \ref{table:uncertain_table} shows statistics collected during the 100 trials in the aerial setup and the underwater setup, showing the average time and the average uncertainty \(\bar{\sigma}_i\) of each target \(i\).

Figure~\ref{fig:uncertaintyepisodes} shows the evolution of the target uncertainties \(\sigma_{i,t}\) during evaluation for both environments and both domains. Each curve corresponds to one of the three targets under either the proposed DSAC policy or the Bug2 baseline. We observe that the DSAC policy resets target uncertainties more frequently and keeps their peaks lower than Bug2, particularly in Env~2 where obstacles induce longer detours. This behaviour is reflected in the lower average uncertainties \(\bar{\sigma}_i\) reported in Table~\ref{table:uncertain_table}. To accelerate convergence, we run DSAC in a distributed fashion with \(K = 5\) parallel simulation workers. Each worker interacts with an independent copy of the environment, collects experience, and shares the replay buffer with the others. The policy and critic networks are shared across workers and updated centrally. 

Overall, we can see that our methodology was able to perform the monitoring of a persistent task for its trained aerial scenario and to generalize and perform it in an underwater setup through a transfer learning methodology. We can also observe that when compared with a behavior-based algorithm, it was able to present better robustness both in terms of task time completion and average uncertainty \(\bar{\sigma}_i\). We highlight the DRL agent's transfer learning capability, where the agent was able to perform the task underwater, having learned only in the aerial domain and presenting a similar overall uncertainty \(\bar{\sigma}_i\) in all targets in the first scenario (27.75, 27.45, 23.84) and the second scenario (26.62, 32.87, 27.72). This shows that HUAUVs have the potential to benefit from DRL, considering a persistent monitoring task. 

The present study uses a 2D abstraction of the HUAUV state and a simplified linear target-uncertainty model with three targets and noiseless target estimates. This formulation allows us to isolate the effect of cross-domain dynamics and range-based sensing on the learned policy, but it does not yet capture several important aspects of real HUAUV deployments, such as full 3D motion, sensor noise, occlusions, and more complex coverage objectives. In particular, the uncertainty dynamics are linear and local, and we do not consider global coverage metrics or long-term temporal-logic constraints.

From the algorithmic perspective, extending the formulation to 3D requires augmenting the state with altitude/depth and extending the action space to include vertical velocity or thrust commands. The DSAC framework and the range-based observation model remain applicable; however, the increased dimensionality will likely require more expressive policy networks and a larger amount of training data. Extending the proposed methodology to 3D hybrid motion, richer uncertainty models, and more realistic monitoring objectives is an important direction for future work.

\section{Conclusions}
\label{conclusion}

In this work, we presented a methodology for persistent monitoring tasks using Hybrid Unmanned Aerial Underwater Vehicles (HUAUVs) operating both aerial and underwater domains. Based on Deep DRL and Transfer Learning, we demonstrated that a single DRL architecture can be effectively trained to perform in two distinct domains, using Lidar for aerial sensing and Sonar for underwater perception. Our results show that the proposed framework is capable of adapting to the different challenges imposed by each environment while maintaining robust monitoring capabilities. These findings indicate that cross-domain learning approaches can provide scalable and efficient solutions for complex monitoring tasks, not only in inspection and mapping but also in search and rescue missions.

Future work will focus on extending the proposed architecture to handle more heterogeneous sensor configurations, integrating additional perception modalities, and testing in real-world HUAUV deployments. Moreover, we plan to explore multi-agent coordination strategies to improve scalability and resilience in large-scale monitoring operations.

\vspace{-2mm}
\section*{Acknowledgment}


The authors would like to thank the VersusAI team. This work was partly supported by the CAPES, FINEP, CNPq, PRH-ANP 22, and the Technological University of Uruguay.

\bibliographystyle{./bibliography/IEEEtran}
\bibliography{./bibliography/IEEEabrv,./bibliography/IEEEexample}

\end{document}